\def\year{2018}\relax
\documentclass[letterpaper]{article} 
\usepackage{aaai18}  
\usepackage{times}  
\usepackage{helvet}  
\usepackage{courier}  
\usepackage{url}  
\usepackage{graphicx}  
\usepackage{amsmath}
\usepackage{dirtytalk}
\usepackage{subcaption}
\usepackage{romannum}
\nocopyright  
\begin{document}
%
\title{Meta-Learning  via Feature-Label Memory Network }
\author{Dawit Mureja \and Hyunsin Park \and Chang D. Yoo\\
Korea Advanced Institute of Science and Technology\\
School of Electrical Engineering\\
\url{{dawitmureja,hs.park,cd_yoo}@kaist.ac.kr}
}
\maketitle
\begin{abstract}
Deep learning typically requires training a very capable architecture using a large dataset. However, many important learning problems demand an ability to draw valid inferences from a small size dataset, and such problems pose a particular challenge for deep learning. In this regard, various researches on \say{meta-learning} are being actively conducted. Recent work has suggested a Memory Augmented Neural Network (MANN) for meta-learning. MANN is an implementation of a Neural Turing Machine (NTM) with the ability to rapidly assimilate new data in its memory, and use this data to make accurate predictions. In models such as the MANN, the input data samples and their appropriate labels from previous step are bound together in the same memory locations. This often leads to memory interference when performing a task as these models have to retrieve a feature of an input from a certain memory location and read only the label information bound to that location. In this paper, we tried to address this issue by presenting a more robust MANN. We revisited the idea of meta-learning and proposed a new memory augmented neural network by explicitly splitting the external memory into feature and label memories. The feature memory is used to store the features of input data samples and the label memory stores their labels. Hence, when predicting the label of a given input, the memory augmented network with separate feature and label memory unit uses the feature memory unit as a reference to extract the stored feature of the input, and based on that feature, it retrieves the label information of the input from the label memory unit.  In order for the network to function in this framework, a new memory-writing module to encode label information into the label memory in accordance with the meta-learning task structure is designed. Here, we demonstrate that the memory-augmented network outperforms MANN by a large margin in supervised one-shot classification tasks using Omniglot and MNIST datasets.  
\end{abstract}

\section{Introduction}
Deep learning is heavily dependent on big data. Traditional gradient based neural networks require extensive and iterative training using large datasets. In these models, training occurs through a continuous update of weight parameters in order to optimize the loss function during training. However, when there is only a little data to learn from, deep learning is prone to poor performance because traditional networks will not acquire enough knowledge about the specific task via weight updates, and hence, they fail to make accurate predictions when tested.
 
\begin{figure*}
	\begin{minipage}{0.4\textwidth}
		\includegraphics[scale= 0.55]{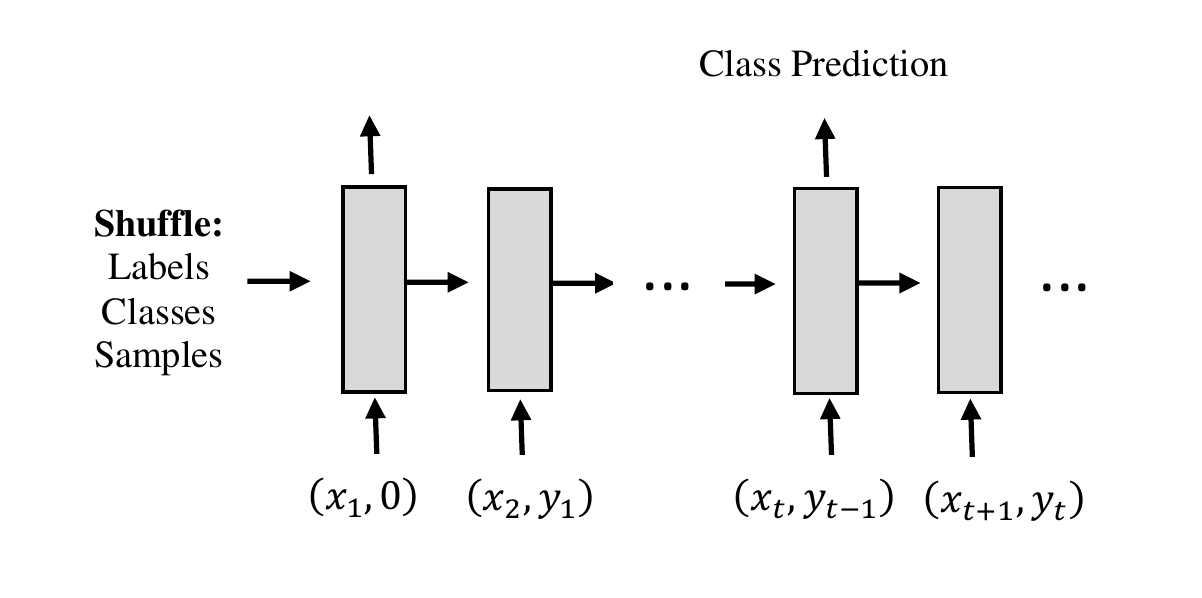}
		\subcaption{Task setup}
	\end{minipage}
	\begin{minipage}{0.6\textwidth}
	 \includegraphics[scale= 0.55]{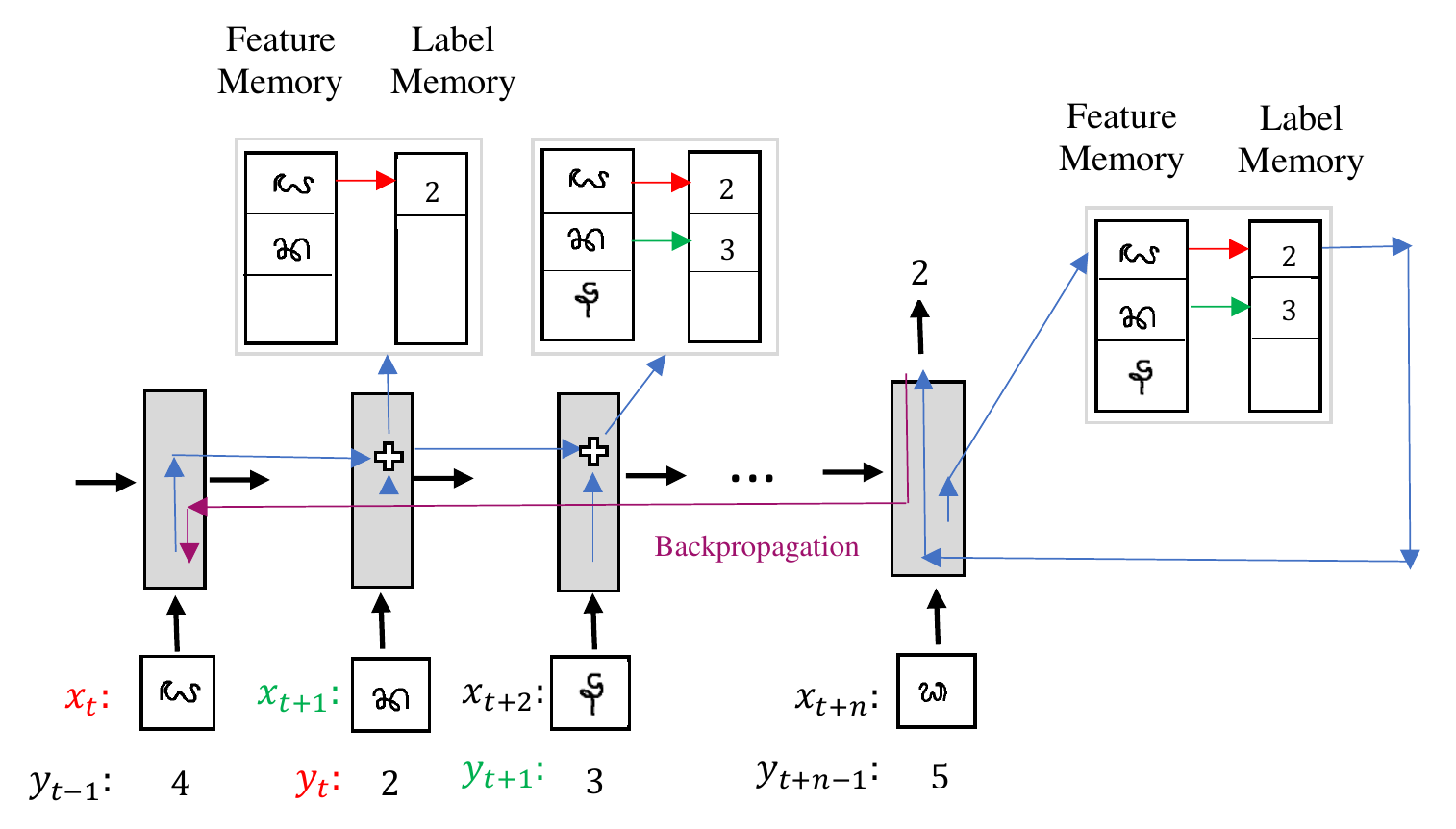}
	 \subcaption{Encoding and Retrieving}
	\end{minipage}
	\caption{Meta-learning task structure. (a) Omniglot images , $\textbf{x}_t$, are presented along with labels in a temporally offset manner. At time step $t$, the network sees an input image $x_t$ with a label $y_{t-1}$ from the previous time step. Labels are also shuffled from episode to episode. This prevents the model from learning sample-class bindings via weight updates, instead it learns to regulate input and output information using its two memories. (b) Here is how the model works. When the network sees an input image for the first time at a certain time step, it stores a particular feature of the input in the feature memory. When the appropriate label is presented at the next time step, the network stores the label information of the input in the label memory. Then, sample-class bindings will be formed between the two memories. When the network is given the same class of image in later time step, the network retrieves the input feature from the feature memory and uses the retrieved information to read the corresponding label memory for prediction.}
	\label{fig:accuracy}
\end{figure*}

Previous works have approached the task of learning from few samples  using different methods such as probabilistic models based on Bayesian learning \cite{bibfile2}, generative models using probability density functions \cite{bibfile3,bibfile9}, Siamese neural networks \cite{bibfile5}, and meta-learning based memory augmented models \cite{bibfile1,bibfile6}.

In this work, we revisited the problem of meta-learning using memory augmented neural networks. Meta-learning is a two-tiered learning framework in which an agent learns  not only about the specific task, for instance, image classification, but also about how the task structure varies across target domains \cite{bibfile7,bibfile1}. Neural architectures with an external memory such as Neural Turing Machines (NTMs) \cite{bibfile8} and memory networks \cite{bibfile11} have shown the ability of meta-learning.

Recent memory augmented neural networks for meta-learning such as MANN \cite{bibfile1} use a plain memory matrix as an external memory. In these models, input data samples and their labels are bound together in the same memory locations.In models such as the MANN, the input data samples and their appropriate labels from previous step are bound together in the same memory locations. This often leads to memory interference when performing a task as they have to retrieve a feature of an input from a certain memory location and read only the label information bound to that location.

Our primary contribution in this work is designing a different version of NTM \cite{bibfile8} by splitting the external memory into feature and label memories to avoid any catastrophic interference. The feature memory is used to store input data features and the label memory is used to encode the label information of the inputs. Therefore, during testing, ideal performance in our model requires using the feature memory as a reference to accurately retrieve the stored feature of an input image and  effectively reading the corresponding label information from the label memory. In order to accomplish this, we designed a new memory writing module based on the meta-learning task structure that monitors the way in which information is written into the label memory.

\section{Related Work}
Our work is based on a recent work done by a \citeauthor{bibfile1} They approached the problem of one shot learning with the notion of meta learning and suggested a Memory Augmented Neural Network (MANN). MANN is an implementation of NTM \cite{bibfile8} with an ability to rapidly assimilate new data, and use this data to make accurate predictions after a few samples. 

In previous implementation of NTM, memory was addressed both by content and location. However, in their work, they presented a new memory access module. This memory access module is called Least Recently Used Access (LRUA)\cite{bibfile1}. It is a pure content-based memory writer that writes memories either to the least recently used location or to the most recently used location of the memory. According to this module, new information is written into rarely used locations (preserving recently encoded information) or it is written to the last used location (to update the memory with newer, and possibly relevant, information).

\section{Task Methodology}
In this work, we used a similar task structure used in recent works \cite{bibfile1,bibfile6}. As we implemented supervised learning, the  model is tasked to infer information from a labelled training data. This involves presenting the label $y_t$ along with input $x_t$ at time step $t$. However, in our work, the training data was presented in the following manner: $ D = \left\{(x_t,y_{t-1})\right\}^T_{t = 1}$, where $D$ is the dataset, $x_t$ is the input at time step $t$ and $y_{t-1}$ is the class label from previous time step $t-1$. Therefore, the model sees the following input sequence: $(x_1,0),(x_2,y_1),...,(x_T,y_{T-1})$ (Figure 1(a)).

Moreover, the label used for a particular class of input images in a certain episode is not necessarily the same as the label used for the same class of input images in another episode. Random shuffling of labels is used from episode to episode in order to prevent the model from slowly learning sample-class bindings in its weights. Instead, it learns to store input information into the feature memory and store the corresponding output information into the label memory, when presented at the next time step, after which sample-class bindings, between the input features in the feature memory and the class labels in the label memory, will be formed for later use (Figure 1(b)).
\section{Memory Augmented Model}
Neural Turing Machine (NTM) \cite{bibfile8} is a memory augmented neural network that has two main components: a controller and an external memory. It can be seen as a differentiable version of a Turing machine. The controller is a neural network that provides an internal representation of the input used by read and write heads to interact with the external memory. It can be either feed-forward or recurrent neural network.

In this work, we designed a memory augmented neural network, a different version of NTM, with its memory split into partitions: Feature memory ($M^f$) and Label memory ($M^l$). The feature memory is used as a reference memory to retrieve the stored representation of an input data. The label memory is used to read an output information of the input based on the retrieved information from the feature memory. In our model, we used Long Short Term Memory (LSTM) \cite{bibfile10} as a controller due to its better performance compared to other controller models. Figure 2 shows the high-level diagram of our model.

\begin{figure}
	\centering
	\includegraphics[scale= 0.6]{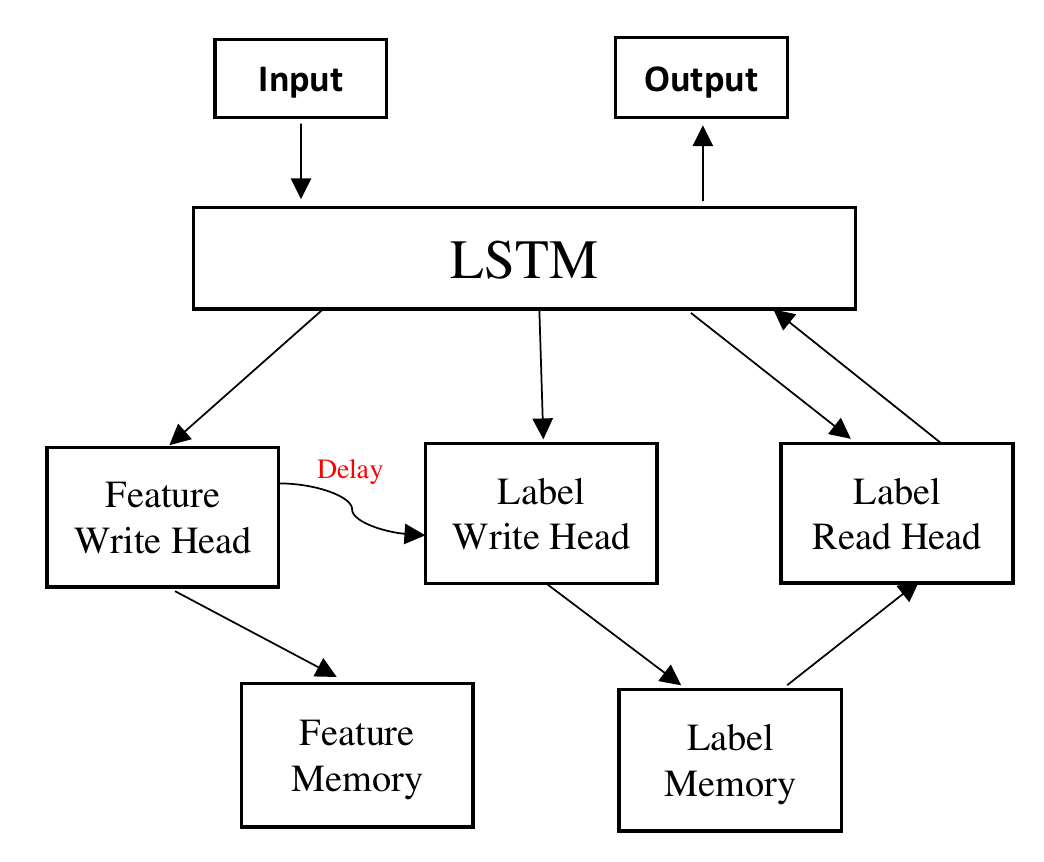}
	\caption{Feature-Label Memory Network (FLMN). It a memory augmented neural network with LSTM controller and two memories (Feature and Label memories). Input features are encoded into the feature memory using the feature memory write head. Labels of the inputs are written into the label memory using the label memory write head. The two write heads are linked recursively in accordance with the task structure. Label read head is used to read label information of a given input from the label memory.}
\end{figure}
Feature-Label Memory Network (FLMN) has two memories, and hence, has two write heads. Feature memory write head writes into the feature memory ($M^f$). Label memory write head is a writer to the label memory ($M^l$). Even though information is encoded in both memories, output information is read only from the label memory using the label read head.

Here is how our model works. Given some input $x_t$ at time step $t$, the controller produces three interface vectors, $k_t$, $a^f_t$ and $a^l_t$. Key vector ($k_t$) is used to retrieve a particular memory, $i$, from a row of the feature memory; i.e. $M^f_t(i)$. Add vectors ($a^f_t$ and $a^l_t$) are used to modify the content of feature memory ($M^f_t$) and label memory ($M^l_t$), respectively. 
\subsection{Reading from the Label Memory}
Before the output information of  the input image $x_t$ is read from the label memory, the corresponding feature of $x_t$ is retrieved from the feature memory using a key $k_t$.
When retrieving memory, the $i^{th}$ row of the feature memory ($M^f_t$) is addressed using \textit{cosine similarity} measure,
\begin{equation}
K[k_t, M^f_t(i)] = \displaystyle \frac{k_{t} \cdot M^f_t(i)}{\|k_t\| \cdot \|M^f_t(i)\|}
\end{equation} 
This measure, $K$, is then used to produce read-weight vector ($w^r_t$) whose elements are computed according to the following softmax:
\begin{equation}
w^r_t(i) \leftarrow \displaystyle \frac{\exp(K[k_t,M^f_t(i)])}{\displaystyle{\sum\limits_{j} \exp(K[k_t,M^f_t(j)])}}
\end{equation}
The read weights are then used to read from label memory ($M^l)$. The read memory, $r_t$, is computed as follows,
\begin{equation}
r_t \leftarrow \sum\limits_{i} w^r_t(i)M^l_t(i)
\end{equation}
\subsection{Writing into the Feature Memory}
In order to write into the feature memory, we implemented the LRUA module \cite{bibfile1} with slight modifications. According to this module, new information is written either into rarely used locations or to the last used location. The distinction between these two options is accomplished by an interpolation using usage weight vector $w^u_t$.

The usage weight vector $w^u_t$ at a given time step is computed by decaying the previous usage weights and adding the current write weights of the feature memory $w^{wf}_t$ and read weights as follows, 
\begin{equation}
w^u_t \leftarrow \gamma w^u_{t-1} + w^{wf}_t + w^r_t
\end{equation}
where, $\gamma$ is a decay parameter.

In order to access the least-used location of the feature memory, least-used weight vector $w^{lu}$ is defined from the usage weight vector $w^u$, 
\begin{equation}
w^{lu}(i) =  \begin{cases} 
1 & $if\ $ w^u(i) = \min(w^u)\\
0 & $otherwise$
\end{cases}
\end{equation}
 Write weights for the feature memory ($w^{wf}$) are then obtained by using  a learnable sigmoid gate parameter to compute a convex combination of the previous read weights and previous least-used weights. 
 \begin{equation}
 w^{wf}_t \leftarrow \sigma(\alpha) w^r_{t-1} + ( 1- \sigma(\alpha))w^{lu}_{t-1}
 \end{equation}
 where, $\displaystyle{\sigma(\alpha) = \frac{1}{1 + \exp(-\alpha)}}$  and $\alpha$ is a scalar gate parameter to interpolate between weights.
 
 Therefore, new content is written either to the previously used memory (if $\sigma(\alpha)$ is 1) or the least-used memory (if $\sigma(\alpha)$ is 0). Before writing into the feature memory, the least used location of  the memory is cleared. This can be done via element-wise multiplication using the least-used weights from the previous time step:
 \begin{equation}
 M^f_t(i) \leftarrow M^f_{t-1}(i) \cdot (1-w^{lu}_{t-1}(i)),\quad \forall{i}
 \end{equation}
 Then writing into memory occurs in accordance with the computed weight vectors using the feature add vector ($a^f_t$) as follows,
 \begin{equation}
  M^f_t(i) \leftarrow M^f_{t}(i) + w^{wf}_t(i)a^f_t,\quad \forall{i}
 \end{equation}

 \subsection{Writing into the Label Memory}
 
 According to (3), the read memory $r_t$ is retrieved from the label memory $M^l_t(i)$ using the read weights $w^r_t$ with the elements computed using (2) which involves the feature memory $M^f_t(i)$. Hence, the label memory should be written in a similar manner as the feature memory so that when an input image  $x_t$ is provided to the network at time step $t$, the network retrieves the stored feature of the input from $M^f_t$ and based on that feature, it extracts the label of the input image from $M^l_t$. 
 
 In order to accomplish the above scenario, we designed a new memory writing module for the label memory. The new module is based on the task setup in which the model was trained. As mentioned earlier, during training, the model sees the following input sequence: $(x_1,0),(x_2,y_1),...,(x_T,y_{T-1})$. The label $y_t$ at time step $t + 1$ is the appropriate label for the input $x_t$ which was presented along with the label $y_{t-1}$ at time step $t$. Based on this observation, we designed a \textbf{recursive memory writing module}.

 According to this module, the label memory write-weight vector $w^{wl}$ at time step $t$ is computed from the previous feature memory write-weight vector $w^{wf}_{t-1}$ in a recursive manner as follows,
 \begin{equation}
  w^{wl}_t(i) \leftarrow w^{wf}_{t-1}(i)
 \end{equation}
 
 The label memory ($M^l$) is then written  according to the write weights  $w^{wl}$ using the label add vector $a^l_t$.
 \begin{equation}
 M^l_t(i) \leftarrow M^l_{t-1}(i) + w^{wl}_t(i)a^l_t,\quad \forall{i}
 \end{equation}
 This memory is then read as shown in (3) to give a read memory, $r$, which will be used by the controller as an input to a softmax classifier, and as an additional input for the next controller state.
 
 Based on this module, the label $y_t$ at time step $t+1$ will be written into the label memory in the same manner as the input $x_t$ (from the previous time step $t$) was written into the feature memory. This enhances the model to accurately retrieve input information from the feature memory and use this feature to effectively read the corresponding output information from the label memory without any interference. 
 
 \section{Experimental Results}
 We tested our model in one-shot image classification tasks using Omniglot and miniMNIST datasets. The omniglot dataset consists of 1623 characters from 50 different alphabets. The number of samples per each class (character) is 20. The dataset is also called MNIST transpose due to the fact that it contains large number of classes with relatively few data samples per class. This makes the dataset ideal for one-shot learning.

\subsection{Experiment Setup}
In this work, we implemented both our model and MANN \cite{bibfile1} and compared their performance in supervised one-shot classification tasks. However, the experimental settings we used for implementing MANN are slightly different from the implementation of MANN in the original paper \cite{bibfile1}.

\begin{figure}[h!]
	\begin{minipage}{0.5\textwidth}
		\includegraphics[scale= 0.6]{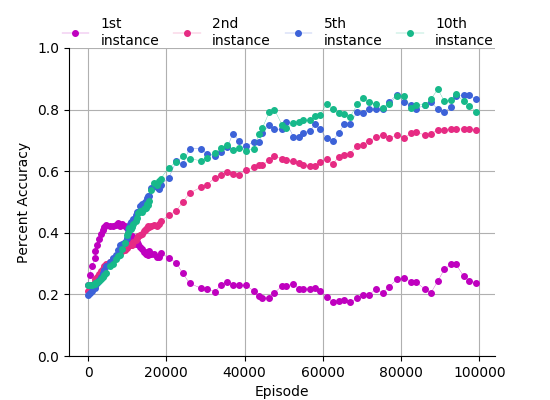}
		\subcaption{Experiment I. Training accuracy for MANN}
	\end{minipage}
	\begin{minipage}{0.5\textwidth}
		\includegraphics[scale= 0.6]{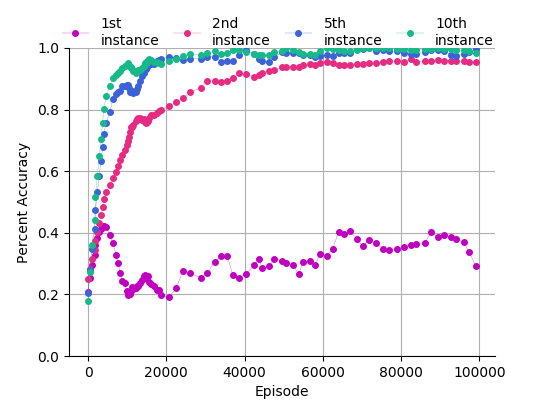}
		\subcaption{Experiment I. Training accuracy for FLMN}
	\end{minipage}
	\caption{Omniglot classification. No data augmentation was performed. In (a) and (b), each episode contains 5 classes and 10 samples per each class. As expected, the $\mathrm{1^{st}}$ instance accuracy is quite low in both models. This is because the models have to do a random guess for the first presentation of the class. However, as we can see from (b), FLMN $\mathrm{1^{st}}$ instance accuracy is more than a blind guess, especially after 20,000 episodes, which indicates that the model is making an educated guess for new classes based on the previous classes it has already seen. For the $\mathrm{2^{nd}}$ and other instances, both models use their memory to achieve better accuracy. The $\mathrm{2^{nd}}$ instance accuracy of FLMN has reached 80\% with in the first 20,000 episodes while the $\mathrm{2^{nd}}$ instance accuracy of MANN reached only to 40\% accuracy.}
\end{figure}

\begin{figure*}
	\begin{minipage}{0.5\textwidth}
		\includegraphics[scale= 0.6]{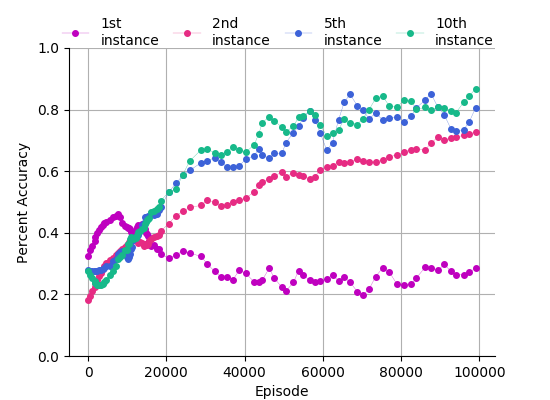}
		\subcaption{Experiment II. Training accuracy for MANN}
	\end{minipage}
	\begin{minipage}{0.5\textwidth}
		\includegraphics[scale= 0.6]{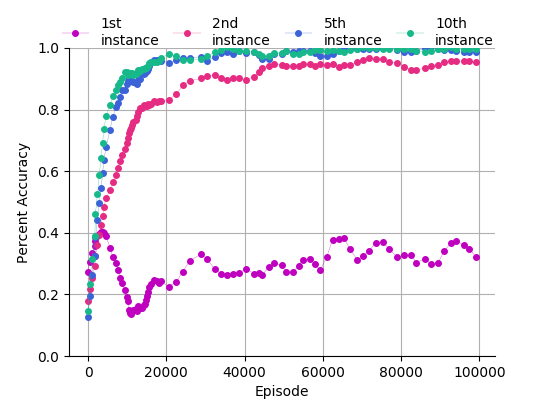}
		\subcaption{Experiment II. Training accuracy for FLMN}
	\end{minipage}
	\caption{Omniglot classification. Data augmentation was performed via rotating and translating random character images in an episode. Each episode contains 5 classes and 10 samples per each class. As we can see from (a) and (b), our model has outperformed MANN by displaying better training accuracies for each instances.}
\end{figure*}

In the paper, the number of reads from the memory used was four. Data augmentation was performed by randomly translating and rotating character images. New classes were also
created through $90^{\circ}$, $180^{\circ}$ and $270^{\circ}$ rotations
of existing data. A minibatch size of 16 was used.

In our case, one read from memory  was used. In order to make a fair comparison, we tried to balance the memory of the two models. we used an $N$ x $M$ memory matrix for MANN, where $N$ is the number
of memory locations, and $M$ is the size at each location. For our model, we split the memory into two and we used $N/2$ x $M$ memory matrix for each memory. Using these settings, we performed three types of experiments.

\subsection{Experiment: Type \Romannum{1}}
 In the first experiment, the original omniglot dataset was used without performing any data modification. Out of the 
 1623 available classes, 1209 classes were used for training and the rest 414 classes were used for testing the models. Note that these two sets are disjoint. Therefore, after training, both models were tested with never-seen omniglot classes. For computational simplicity, image sizes were down scaled to $20\times20$. One-hot vector representations were used for class labels and training was done using 100,000 episodes. Several experiments were performed for different number of classes (and different number of samples per each class) in an episode. Figure 3 shows the training accuracy of the models for 5 classes and 10 samples (per each class) in an episode. 

As we can see from Figure 3, our model has outperformed MANN in making accurate predictions. The $\mathrm{2^{nd}}$ instance accuracy of our model has reached nearly $\mathrm{80\%}$ accuracy within the first 20,000 episodes of training, while the $\mathrm{10^{th}}$ instance accuracy of MANN could only reach $\mathrm{60\%}$ accuracy.

\subsection{Experiment: Type \Romannum{2}}

In our second experiment, we performed data augmentation without creating new classes. The dataset was augmented by rotating and translating random character images of an episode. The angle for rotation was chosen randomly from a uniform distribution $\left[\frac{-\pi}{16},\frac{\pi}{16}\right]$ with a size of an episode. This was accompanied by a translation in the x and y dimensions with values uniformly sampled between -10 and 10 pixels. Images were then downscaled to $20\times20$

In a similar manner as the previous experiment, 1209 classes (plus augmentations) were used for training and 414 classes (plus augmentations) were used for testing. Figure 4 shows the training accuracy of MANN and our model for 100,000 episodes.

Not only has our model performed better in making accurate predictions but also has learned faster than MANN. This can be shown by plotting the \textit{loss graph} of training for the two experiments (Figure 5). 

\begin{figure*}[h!]
	\begin{minipage}{0.5\textwidth}
		\includegraphics[scale= 0.6]{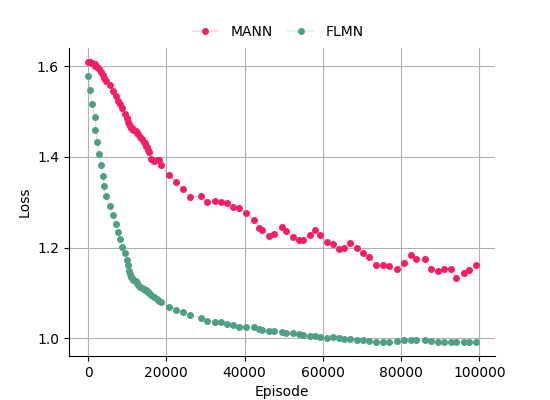}
		\subcaption{Experiment I. Loss graph for MANN and FLMN}
	\end{minipage}
	\begin{minipage}{0.5\textwidth}
		\includegraphics[scale= 0.6]{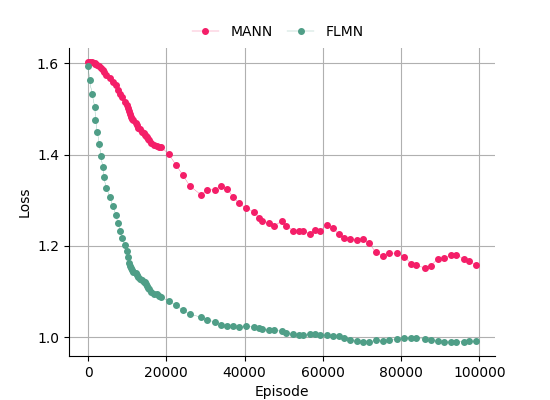}
		\subcaption{Experiment II. Loss graph for MANN and FLMN}
	\end{minipage}
	\caption{Loss graph. For the same learning rate, the loss graph of FLMN has fallen sharply compared to the loss graph of MANN in both (a) and (b). This indicates that FLMN has learned the task of one-shot classification faster than MANN and explains why it has demonstrated higher training accuracy within the first few episodes.}
\end{figure*}

In both types of experiments, the training process  has stopped at the mark of the 100,000 episode. Without any further training, the models were tested with never-seen omniglot classes from the testing set. The testing results are summarized in the Table 1. We borrowed the test result of MANN from \citeauthor{bibfile1} for a reference. 

As we can see from the table, our model has demonstrated higher classification accuracy in both experiments compared to MANN. FLMN has reached an accuracy of 85.6\% (Experiment I) and 86.5\% (Experiment II) on just second presentation of an input sample from a class with in an episode reaching up to 94.1\% and 94.4\% accuracy by the $\mathrm{10^{th}}$ instance, respectively. On the other hand, MANN achieved an accuracy of 66.7\% (Experiment I) and 65.5\% (Experiment II) in the $\mathrm{2^{nd}}$ instance reaching up to 78.1\% and 77.2\% accuracy by the $\mathrm{10^{th}}$ instance, respectively.

\begin{table*}
	\centering
	\caption{Test-set classification accuracies of MANN and FLMN for Experiment I and Experiment II}
	\begin{tabular}{c|c c c c c c} \hline
		\multicolumn{1}{c|}{} &\multicolumn{6}{c}{INSTANCE (\% CORRECT)}\\
		Model & $\mathrm{1^{st}}$ & $\mathrm{2^{nd}}$ & $\mathrm{3^{rd}}$ & $\mathrm{4^{th}}$ & $\mathrm{5^{th}}$ & $\mathrm{10^{th}}$ \\ \hline
		MANN \cite{bibfile1}  & 36.4 & 82.8 & 91.0 & 92.6 & 94.9 & 98.1\\ \hline
		MANN (Experiment I) & 21.6 & 66.7 & 74.0 & 76.0 & 76.3 & 78.1\\
		\textbf{FLMN} (Experiment I) & \textbf{31.1} &\textbf{85.6} & \textbf{88.7}&\textbf{89.5} &\textbf{91.0} &\textbf{94.1}\\ \hline
		MANN (Experiment II)  & 22.2 & 65.5 & 72.0 & 74.5 & 76.4 & 77.2\\ 
		\textbf{FLMN} (Experiment II) & \textbf{33.9} &\textbf{86.5} & \textbf{89.1}&\textbf{89.7} &\textbf{91.1} &\textbf{94.4}\\
		\hline
		
	\end{tabular}
	
	\label{tab:data1}
\end{table*}

\subsection{Zero-shot learning}
In this experiment, the models were tasked to perform MNIST classification after being trained with omniglot dataset. We used 1209 classes of omniglot dataset for training. For testing, we prepared a miniMNIST dataset. miniMNIST contains only 20 image samples per each class which are randomly selected from the original MNIST dataset. The images were downscaled to $20\times20$. After 100,000 episodes of training, the models were tested with never-seen MNIST classes. Testing results are summarized in the following table.

\begin{table}[h]
	\caption{Test-set classification accuracies of MANN and FLMN for zero-shot learning}
	\centering
	\begin{tabular}{c|c c c} \hline
		\multicolumn{1}{c|}{} & \multicolumn{3}{c}{INSTANCE (\% CORRECT)}\\
		Model & $\mathrm{1^{st}}$ & $\mathrm{2^{nd}}$ & $\mathrm{10^{th}}$ \\ \hline
		MANN & 14.6 & 37.3 & 52.0 \\
		\textbf{FLMN} & \textbf{28.5} &\textbf{67.6} & \textbf{80.5}\\
		\hline
	\end{tabular}
\end{table}

As we can refer from Table 2, FLMN was able to achieve $80.5\%$ accuracy on the $10^\mathrm{th}$ instance in classifying never-seen-before images from miniMNIST dataset after being trained with omniglot dataset. 
\section{Conclusion}
In this paper, we implemented meta-learning framework and proposed Feature-Label Memory Network (FLMN). The novelty of our model is that it stores input data samples and their matching labels into separate memories preventing any memory interference. We also introduced a new memory writing method associated with the task structure of meta-learning. We have shown that our model has outperformed  MANN in supervised one-shot classification tasks using Omnigot and miniMNIST datasets. Future work includes testing our model with more complex datasets and experimenting the performance of our model in other tasks.
\nocite{bibfile4,bibfile12}
\bibliography{paper}
\bibliographystyle{aaai}
\end{document}